\documentclass{article}
\usepackage{spconf}
\usepackage{amsmath}
\usepackage{graphicx}

\usepackage{subfig}
\usepackage{multirow}

\title{Light Robust Monocular Depth Estimation For Outdoor Environment Via Monochrome And Color Camera Fusion}
\name{Hyeonsoo Jang, Yeongmin Ko, Younkwan Lee, and Moongu Jeon* \thanks{Corresponding author: Moongu Jeon, mgjeon@gist.ac.kr}}
\address{School of Electrical Engineering and Computer Science \\ Gwangju Institute of Science and Technology, Korea}
\begin{document}
%
\maketitle
\begin{abstract} 

\end{abstract}
Depth estimation plays a important role in SLAM, odometry, and autonomous driving. Especially, monocular depth estimation is profitable technology because of its low cost, memory, and computation. However, it is not a sufficiently predicting depth map due to a camera often failing to get a clean image because of light conditions. To solve this problem, various sensor fusion method has been proposed. Even though it is a powerful method, sensor fusion requires expensive sensors, additional memory, and high computational performance.

In this paper, we present color image and monochrome image pixel-level fusion and stereo matching with partially enhanced correlation coefficient maximization. Our methods not only outperform the state-of-the-art works across all metrics but also efficient in terms of cost, memory, and computation. We also validate the effectiveness of our design with an ablation study.

\section{Introduction}
\label{sec:intro}
Depth estimation is a key technique of 3D reconstruction, SLAM, visual odometry, and autonomous driving. Monocular depth estimation, which uses only one camera to predict depth, has advantages of low cost, memory, and computation efficiency. However, estimating a depth map from only one camera is challenging because of an ill-pose problem and defects of the image sensor itself. Since Eigen et al. \cite{eigen_kitti} present CNN-based monocular depth estimation, significant improvements have been made and the state-of-the-art works show a reasonable depth map that overcomes the ill-posed problem \cite{review_monocular2, kim2016unified, Lainaetal, Liuetal}.

A color camera often fails to get clean images because of light smudge, reflection, or insufficient brightness \cite{camera_limit}. Therefore, it is a challenge to get an accurate dense depth map, especially outdoor scenes. To address these problems, sensor fusion that complements the drawbacks of sensors or multi-spectral imaging methods has been proposed using LiDAR, Radar, or multi-spectral camera \cite{review_sensorfusion}. However, sensor fusion suffers from considerable memory, heavy computation, and expansive sensor cost. Furthermore, multiple sensors must be well-calibrated and synchronized to get accurate data.

\begin{figure}[tb]
  \centering 
  \subfloat[RGB image]{{\includegraphics[width=4.2cm]{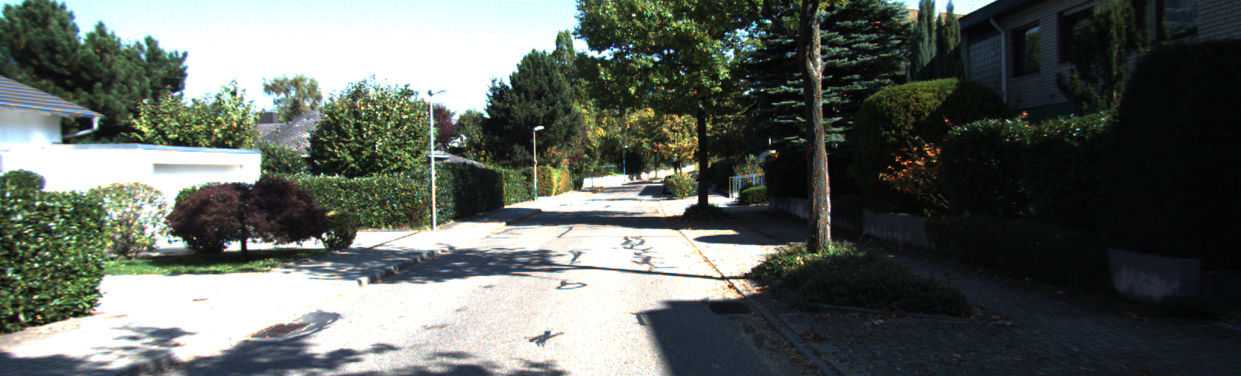}}}
  \hfill
  \subfloat[Monochrome image]{{\includegraphics[width=4.2cm]{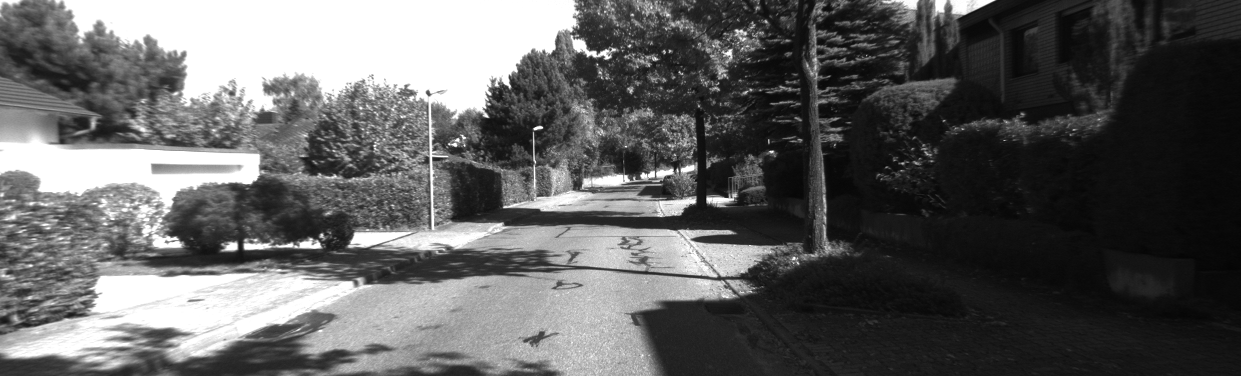}}}
  \hfill
  \subfloat[Lightness channel]{{\includegraphics[width=4.2cm]{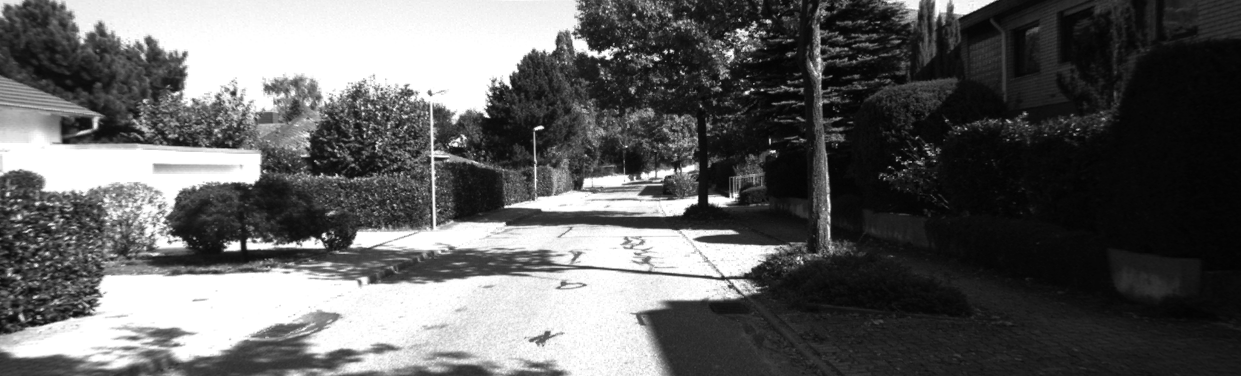}}}
  \hfill
  \subfloat[Fusion image]{{\includegraphics[width=4.2cm]{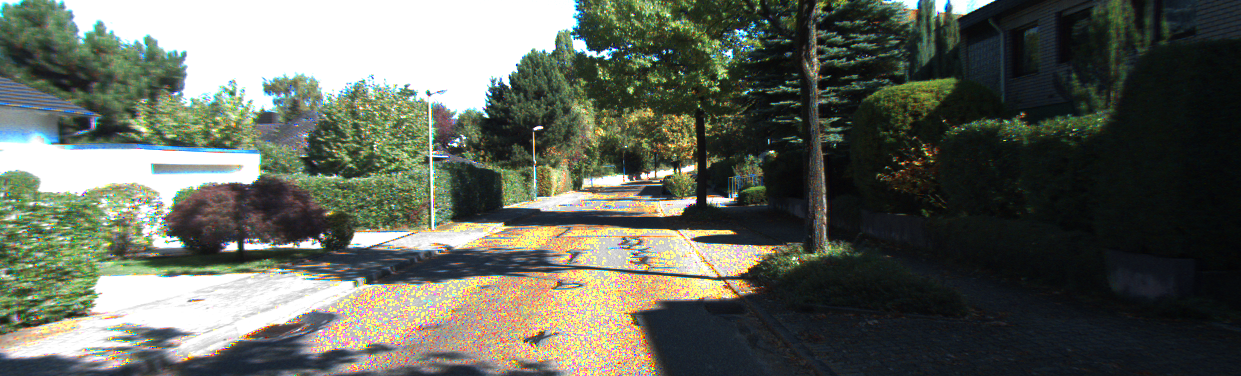}}}

  \caption{Sample of RGB, monochrome, Lightness, Fusion image. The RGB image and its lightness channel are difficult to identify the detail of the road and side for reflection. While the monochrome and the fusion image are more distinguishable}
  \label{fig:3}
\end{figure}

We propose the RGB and monochrome image fusion method to acquire a depth map with encoder-decoder networks. A color camera sensor receives photons and then separates them into red, green, blue by using a Bayer filter. In this process, the recognized spectrum and quantum efficiency are reduced as well as images are corrupted by filter noise, which is inferior to the image obtained by monochrome camera \cite{MonochromeandColor}. Thus, if using both monochrome and color camera, then there is an effect similar to sensor fusion in that it complements flaws of the color camera. Moreover, a monochrome camera is inexpensive and does not require considerable memory and computation. We convert an RGB image to HLS image and replace the lightness channel with the monochrome image. At that point, to reduce the disparity between the two images, the monochrome image is divided and only the bottom parts are warped with enhanced correlation coefficient maximization \cite{ECC}.

To the best of our knowledge, this is the first approach to use monochrome and color images in network base monocular depth estimation. Although a monochrome camera has the same limitations as a color camera sensor, it is worth clarifying the benefits and limitations.

The contributions of our work are:
\begin{itemize}
\item We introduce a monochrome and color image fusion method for reducing the influence of light to enhance the accuracy of depth prediction with advantages of low-cost computation and memory.
\item We design a method independent of depth prediction networks which is why it is applicable to any architecture. 
\item We demonstrate the effectiveness of our proposed method and it improves accuracy significantly compared with the state-of-the-art methods.
\end{itemize}

\section{Proposed Method}
\label{sec:method}
This section introduces the monochrome and color image fusion method. We adopt pixel-level fusion because we assume that a monochrome image is superior to a color image in all pixels so as to utilize most of it. Due to the disparity between monochrome and color images, we conduct an image alignment by enhanced correlation coefficient(ECC) maximization\cite{ECC} to warp the image. An overview of the method is shown in Fig 2 and the resulting image is shown in Fig 1.

In general, since a monochrome image is less affected by light and has less noise than a color image, The former is better able to distinguish objects than the latter. However, it is inadequate to use only monochrome images. First of all, the monochrome image does not contain color information, which is valuable information for visual estimation. Second, traditional backbone networks and pretrained weights are based on an RGB image, which means they are optimized and obtained the best results when using RGB color space.

\begin{figure}[tb]
  \centering 
  \includegraphics[width=8.5cm]{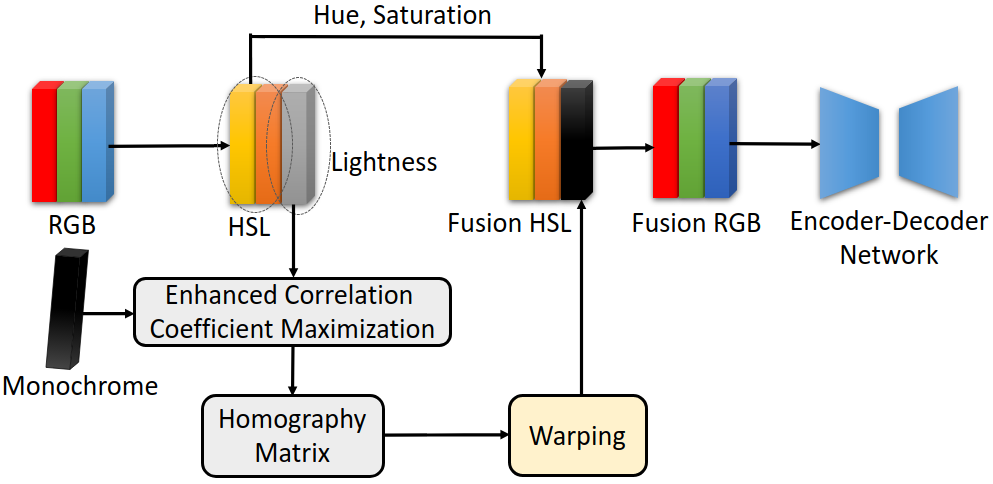}
  \caption{Overview of the proposed fusion method.}%

  \label{method}
\end{figure}

\subsection{Monochrome and color images fusion}

Color can divide into various components depending on the attributes such as hue, brightness, lightness, chroma, and saturation\cite{ColorSpace}. Among them, the brightness and lightness channels have the most similar characteristics to the monochrome image and include noise derived from the light effect. Brightness means a value of light and lightness is relative to white in the scene. We choose HSL color space for the sake of using lightness instead of brightness channel. In HSL color space, color become pure white when lightness increases and approaches black as it decreases regardless of hue and saturation. This reflects the addition of light in the real outdoor environment. RGB images were converted into HSL space images and separated the lightness channel.

\subsection{Image alignment}

We reduced the disparity caused by the distance between the lightness image and the monochrome image partially through an enhanced correlation coefficient(ECC) algorithm \cite{ECC}. Since the two cameras project the almost same scene in 3D into 2D, it can be modeled with a homography matrix.

Applying ECC to the entire image requires heavy computation and the result is inaccurate. Fundamentally, the homography matrix cannot completely present the 3D changes. In particular, if the range of depth is wide in a single image, the calculated homography matrix is far from the perfect warping matrix. To solve this problem, we divide an image into 25 equal parts. As the distance from the camera increases, the disparity of the two images decreases. Therefore, pixels that are long-distance from the camera in the image do not need to be fixed. In the outdoor scene, generally, the bottom of the image is close to the camera. We applied the ECC algorithm only five parts of the bottom of the divided image. By doing that, we were able to not only increase the accuracy of the homography matrix but also reduce the computational cost. The monochrome images are warped using a homography matrix and empty parts are replaced by replicating the original image. Finally, the warped image replaces the lightness channel of the existing color image and converts fusion HSL image into RGB color space.

Traditional encoder-decoder depth estimation networks apply pretrained data trained by ImageNet\cite{ImageNet} to improve performance and avoid overfitting. Since these architectures and weights are trained with RGB images, converting from HSL to RGB color space takes advantage of pretrained weights. The fusion images are used as input batch for the depth estimation network.

\section{Experiments}
\label{sec:pagestyle}

\begin{table*}[htbp]
\caption{Comparison of performances on the KITTI (Eigen split). The measurement distance ranges from 0m to 80m. The best results are in bold and the second-best results are underlined.}
\begin{center}
\begin{tabular}{c|cccc|ccc}
\hline
\multirow{2}{*}{Method}                  & \multicolumn{4}{c|}{Lower value is better} & \multicolumn{3}{c}{Higher value is better}          \\ \cline{2-8} 
                                         & Abs Rel   & Sq Rel   & RMSE    & RMSE log  & $\delta < 1.25$ & $\delta < 1.25^{2}$ & $\delta< 1.25^{3}$ \\ 
\hline
Eigen et al. \cite{eigen_kitti}          & 0.203     & 1.548    & 6.307   & 0.282     & 0.702                & 0.898                & 0.967  \\
Liu et al. \cite{Liuetal}                & 0.201     & 1.584    & 6.471   & 0.273     & 0.680                & 0.898                & 0.967  \\
Godard et al. \cite{monodepth1}          & 0.114     & 0.898    & 4.935   & 0.206     & 0.861                & 0.949                & 0.976  \\
Kuznietsov et al. \cite{Kuznietsovetal}  & 0.113     & 0.741    & 4.621   & 0.189     & 0.862                & 0.960                & 0.986  \\
Gan et al. \cite{Ganetal}                & 0.098     & 0.666    & 3.933   & 0.173     & 0.890                & 0.964                & 0.985  \\
Fu et al. \cite{Fuetal}                  & 0.072     & 0.307    & 2.727   & 0.120     & 0.932                & 0.984                & 0.994  \\
Yin et al. \cite{Yinetal}                & 0.072     & -        & 3.258   & 0.117     & 0.938                & 0.990                & 0.998  \\
BTS \cite{BTS}                           & 0.059     & 0.245    & 2.756   & 0.096     & 0.956                & 0.993                & 0.998  \\
DPT-Hybrid \cite{DPT}                    & 0.062     & -        & 2.573   & 0.092     & 0.959                & \underline{0.995}&\textbf{0.999}  \\
Adabins \cite{Adabins}                   & \underline{0.058} & \underline{0.190}& \underline{2.360} & \underline{0.088}& \underline{0.964} & \underline{0.995}&\textbf{0.999}  \\ 
\hline
\textbf{proposed + BTS}                  & \underline{0.058} & 0.206   & 2.444     & 0.089    & 0.961        & \underline{0.995}    &\textbf{0.999} \\
\textbf{proposed + Adabins}              &\textbf{0.052} &\textbf{0.167} &\textbf{2.277}&\textbf{0.080}&\textbf{0.974}&\textbf{0.997}&\textbf{0.999} \\
\hline 
\end{tabular}
\label{result tab1e}
\end{center}
\end{table*}

\begin{table}[htbp]
\caption{Ablation study results with Adabins baseline. F(Fusion): Replacing the brightness channel with monochrome image, Warp: Applying warping algorithm, Part: Dividing an image into 25 equal parts and applying warping only bottom parts}
\begin{tabular}{l|cccc}
\hline
\multicolumn{1}{c|}{Variant} & Sq Rel & RMSE  & $\delta < 1.25$ & $\delta < 1.25^{2}$             \\
\hline
F                            & 0.174           & 2.322          & \textbf{0.973}   & 0.996          \\
F+Warp                       & 0.216           & 2.500          & 0.958            & 0.994          \\
\textbf{F+Warp+Part}         & \textbf{0.167}  & \textbf{2.277} & \textbf{0.973}   & \textbf{0.997} \\ \hline
\end{tabular}
\end{table}

We train the fusion image with the state-of-the-art monocular depth estimation network. We adapt BTS\cite{BTS} and Adabins\cite{Adabins} as a baseline models. The performance of our method compares with other previous studies as well as the results of the original baseline model.

\subsection{KITTI Datasets}

KITTI provides the dataset with stereo images and corresponding 3D LiDAR data of outdoor scenes from "city", residential", "road", and "campus" captured using equipment mounted on a vehicle \cite{KITTI}. Particularly, it provides both RGB and monochrome images, which the same resolution of around 1241 x 376. We follow the split proposed by Eigen et al. \cite{eigen_kitti}. The subset from the left view images which is about 26,000 images adopted to training and 697 images are used for evaluation. The depth maps for a single image have an upper bound of 80 meters. We use a random crop of size $704 \times 352$ for training and crop as defined by Garg et al. \cite{Gargetal} for evaluation. Additionally, predicted depth maps are bilinearly upsampled to match the ground truth resolution.

\subsection{Implementation details}

We implement the proposed fusion method using OpenCV\cite{OpenCV} and CNN and transformer networks in PyTorch. We iterate the ECC algorithm 20 times on five image blocks to obtain a homography matrix. For training, we follow original BTS \cite{BTS} and AdaBins \cite{Adabins} optimizer, detail parameters, and backbone architecture. For BTS \cite{BTS} based model, we use Adam optimizer \cite{Adam} with $\beta_{1} = 0.9$, $\beta_{2} = 0.99$, and $\epsilon=10^{-6}$, learning rate decay polynomially from $10^{-6}$ with power 0.9 for training, and choose ResNeXt-101 \cite{ResNeXt} with pretrained weights using ImageNet \cite{ImageNet} because it shows best accuracy on KITTI dataset and fix parameters of first two layers for these layers are trained well to extract low-level features \cite{BTS}.  AdamW optimizer \cite{AdamW} with weight-decay $10^{-2}$ and pretrained EfficientNet-B5 \cite{EfficientNet} is chosen for Adabins \cite{Adabins} based model and apply the 1-cycle policy \cite{superconvergence} for the learning rate with $max \_ lr = 3.5\times10^{-4}$. For the first 30\% of iterations apply linear warm-up from $max \_ lr / 25$ to $max \_ lr$. Afterwards, it follows cosine annealing to $max \_ lr / 75$ \cite{Adabins}. The total number of epochs is set to 50 and batch size 16 for both the BTS base and Adabins base model. We use two NVIDIA GeForce RTX 3090 GPUs for all experiments.

To avoid the overfitting problem, online image augmentation is conducted after the fusion process. We use random horizontal flipping as well as random crop. We also use contrast, brightness, a color adjustment in a range of [0.9, 1.1], with 50\% chance and random rotation in the range of [-1, 1] and [-2.5, 2.5] degrees.

\begin{figure*}[htbp]
  \begin{minipage}[c]{18cm}
    \centering
    \raisebox{17 pt}{\rotatebox[origin=t]{90}{RGB}}
    \includegraphics[width=5.6cm]{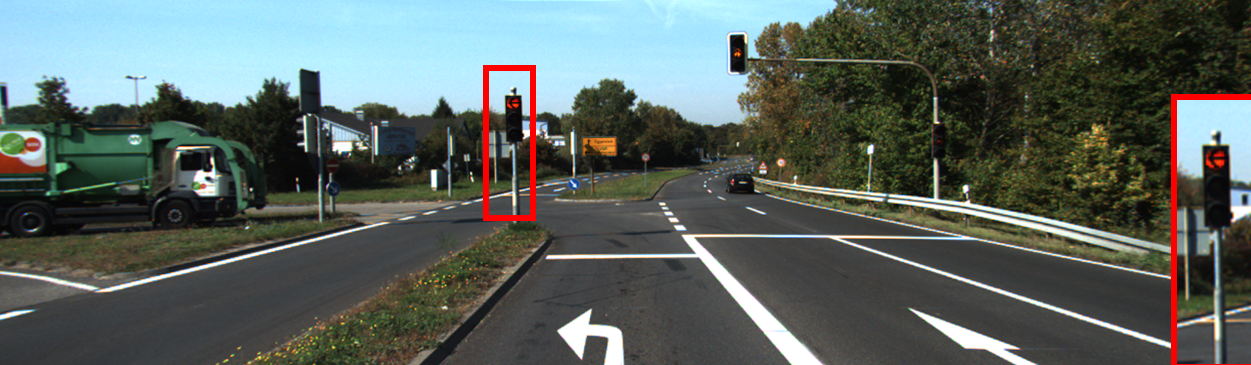}
    \includegraphics[width=5.6cm]{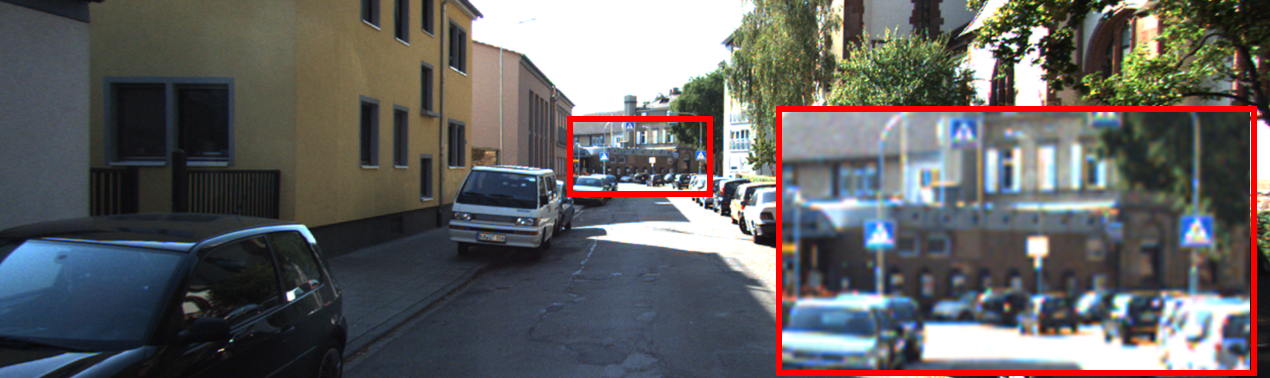}
    \includegraphics[width=5.6cm]{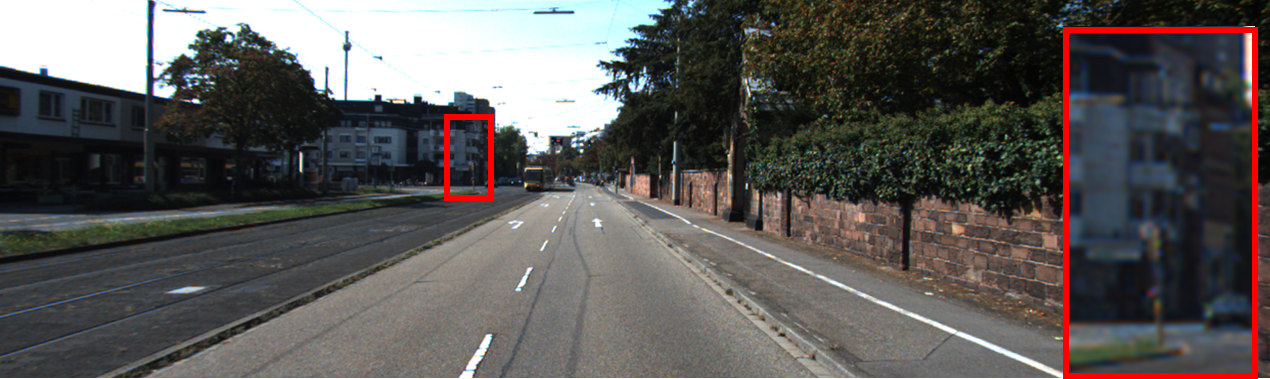}
  \end{minipage}
  \begin{minipage}[c]{18cm}
    \centering
    \raisebox{17 pt}{\rotatebox[origin=t]{90}{BTS}}
    \includegraphics[width=5.6cm]{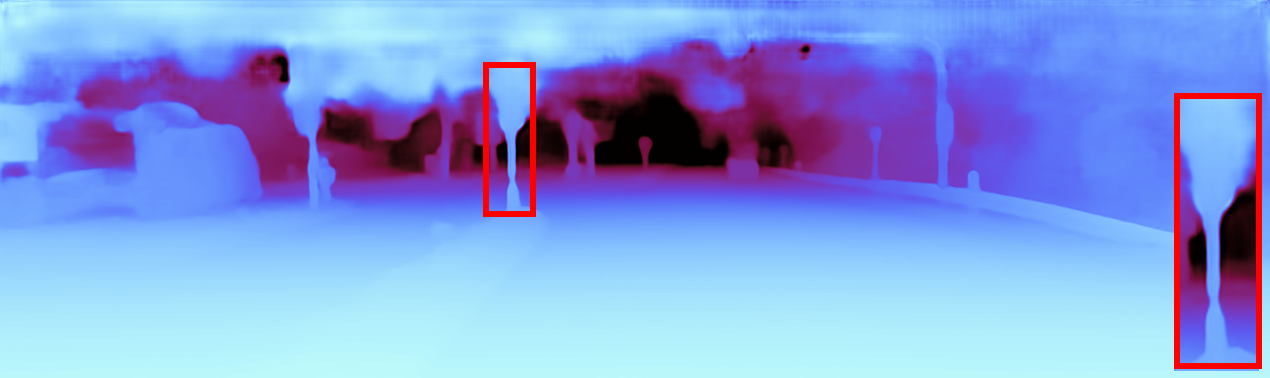}
    \includegraphics[width=5.6cm]{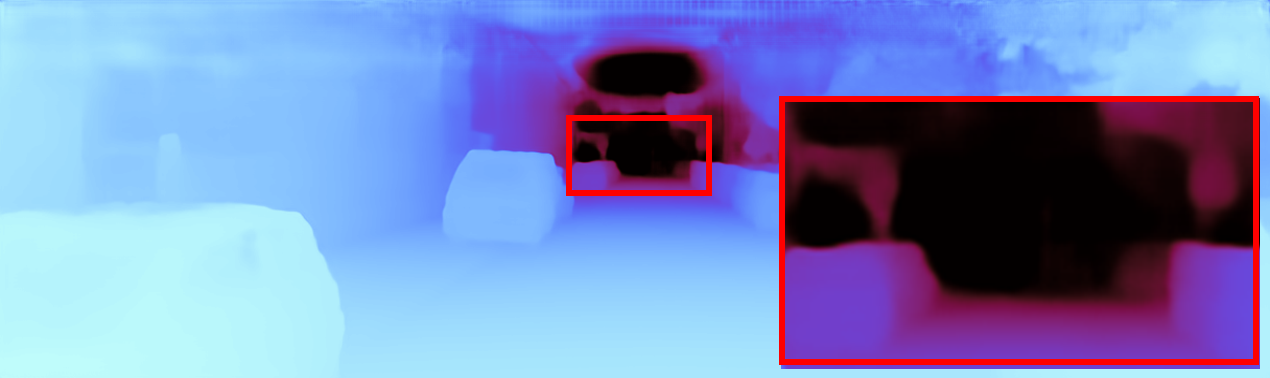}
    \includegraphics[width=5.6cm]{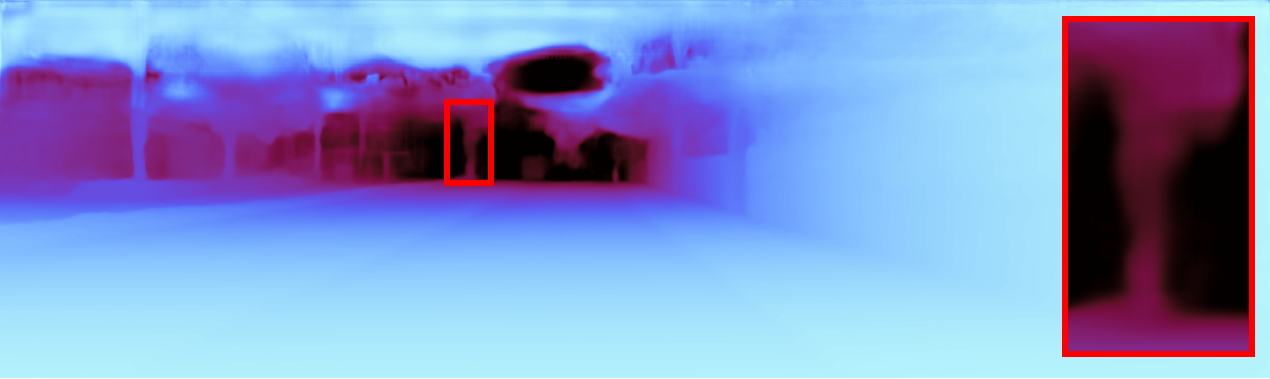}
  \end{minipage}
  \begin{minipage}[c]{18cm}
    \centering
    \raisebox{17 pt}{\rotatebox[origin=t]{90}{Ours(BTS)}}
    \includegraphics[width=5.6cm]{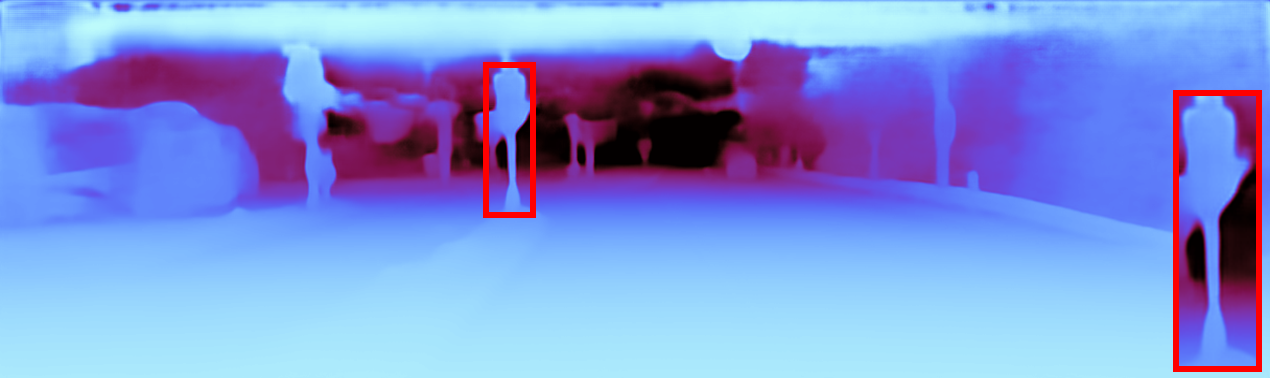}
    \includegraphics[width=5.6cm]{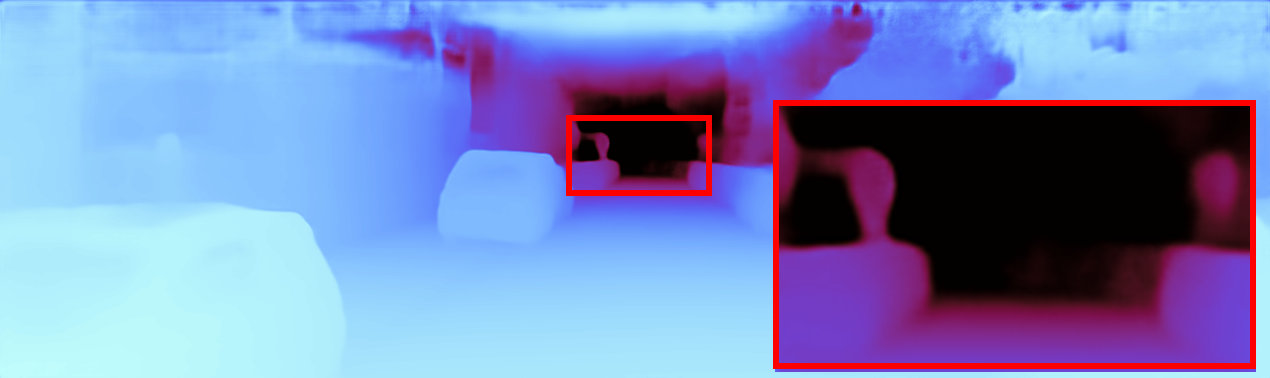}
    \includegraphics[width=5.6cm]{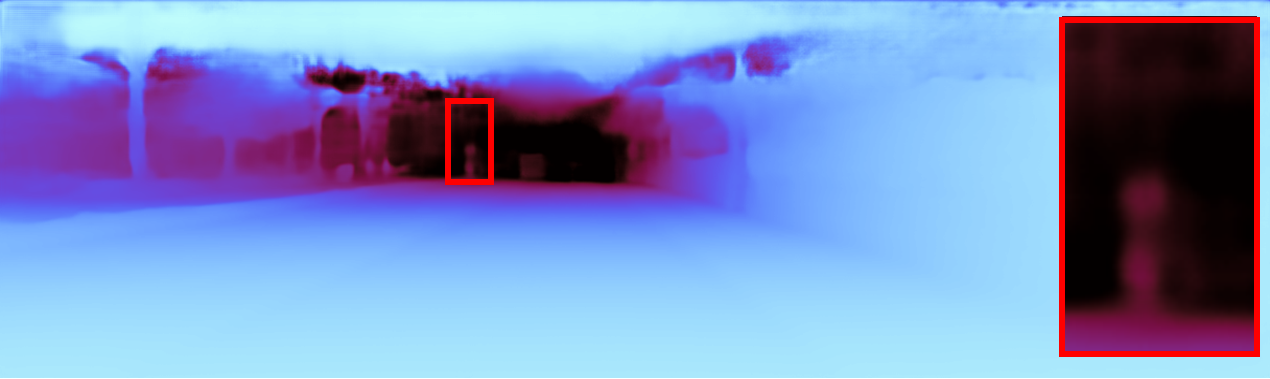}
  \end{minipage}
  \begin{minipage}[c]{18cm}
    \centering
    \raisebox{17 pt}{\rotatebox[origin=t]{90}{Adabins}}
    \includegraphics[width=5.6cm]{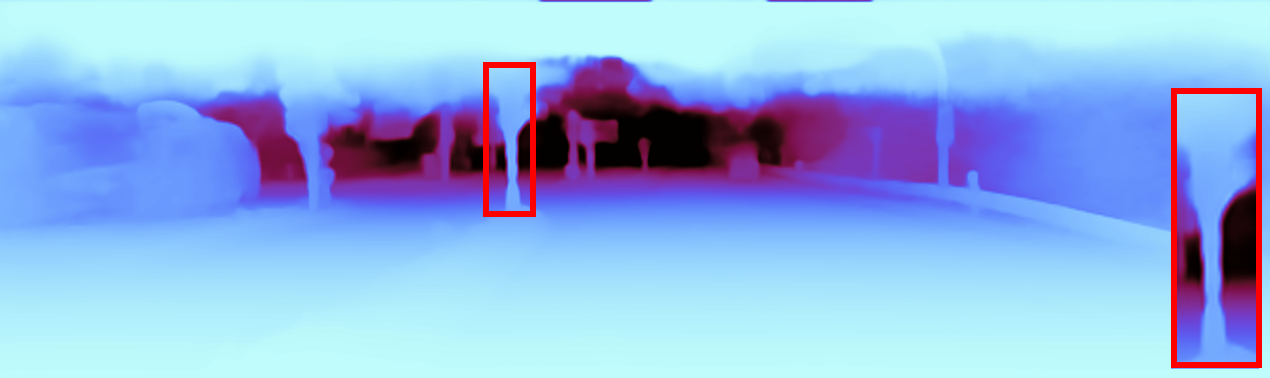}
    \includegraphics[width=5.6cm]{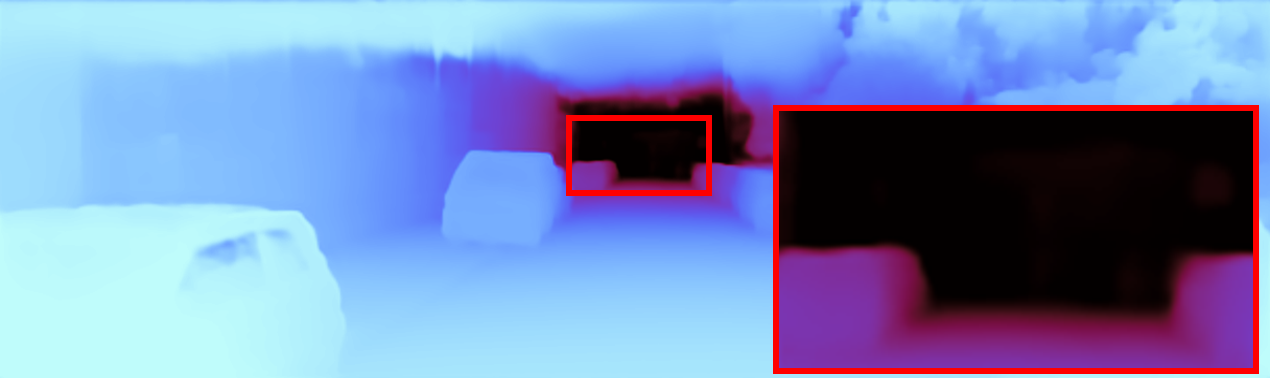}
    \includegraphics[width=5.6cm]{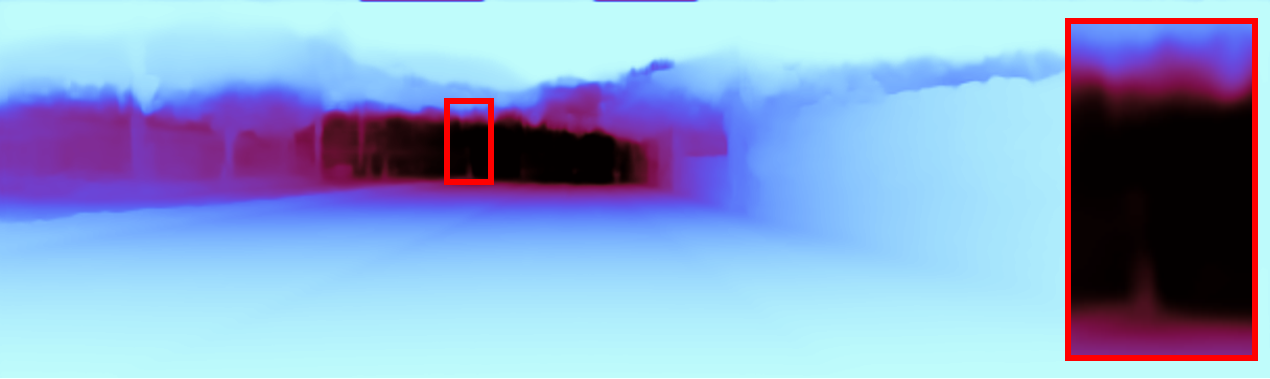}
  \end{minipage}
  \begin{minipage}[c]{18cm}
    \centering
    \raisebox{17 pt}{\rotatebox[origin=t]{90}{Ours(Adabins)}}
    \includegraphics[width=5.6cm]{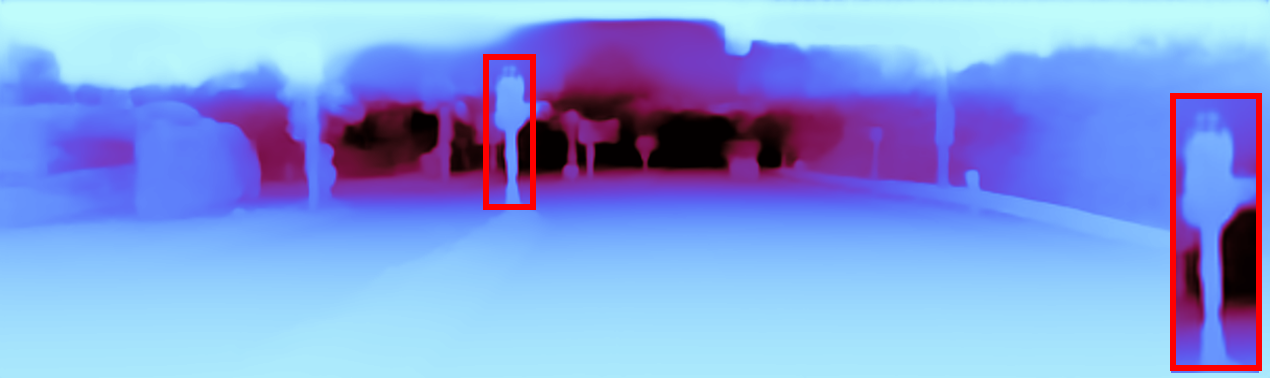}
    \includegraphics[width=5.6cm]{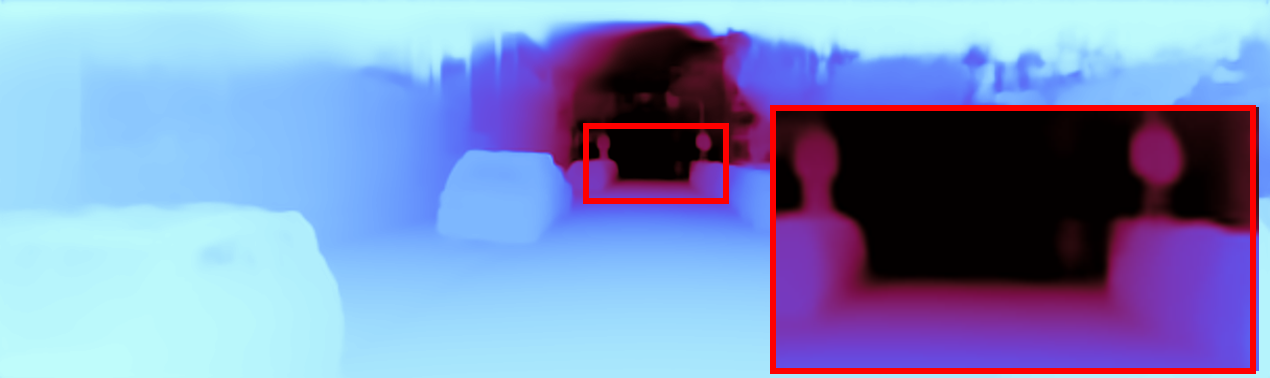}
    \includegraphics[width=5.6cm]{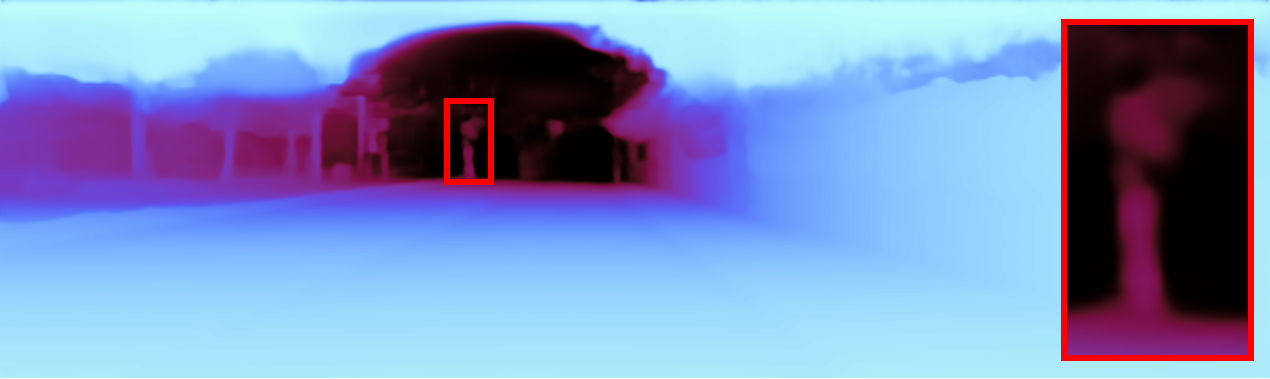}
  \end{minipage}
  
  \caption{Qualitative comparison with other methods on the KITTI Eigen test split. 1st rows: color images. 2nd rows: BTS, 3rd rows: Ours(BTS base), 4th rows: Adabins, 5th rows: Ours(Adabins base)}%
  \label{fig:4}
\end{figure*}

\subsection{Evaluation results}

We use the following metrics used in prior work \cite{eigen_kitti} to compare accuracy.
\\

Abs Rel: $\frac{1}{\left| T \right|}\sum_{y\in T}^{}\frac{\left|y-\hat{y} \right|}{y}$

Sq Rel: $\frac{1}{\left| T \right|}\sum_{y\in T}^{}\frac{\left||y-\hat{y} \right||^{2}}{y}$

RMSE: $\sqrt{\frac{1}{\left| T \right|}\sum_{y\in T}^{}(y-\hat{y})^{2}}$

RMSElog: $\sqrt{\frac{1}{\left| T \right|}\sum_{y\in T}^{}(\log{y-\log{\hat{y})^{2}}}}$

Threshold: \% of $y$  s.t. $max(\frac{y}{\hat{y}}, \frac{\hat{y}}{y}) = \delta < thr \;\text{for}\; thr =  1.25, 1.25^{2}, 1.25^{3};$ where $T$ denotes a total number of valid pixels in the ground truth. $y$ is a ground truth pixel, $\hat{y}$ is a pixel in the predicted depth image.

As detailed in table 1, Our method outperforms the state-of-the-art works with a significant margin on all metrics regardless of a model. As shown in Fig 3, our method recognizes well distant objects regardless of low light or reflection. For our ablation study, we verify the influence of each design as shown in table 2.



\section{Conclusion and Discussion}
We introduced a monochrome and color image fusion method for CNN-based monocular depth estimation. Our experiments show significant improvements over original based models, especially less affected by light and recognized long distant objects well.

Although our method was achieved excellent results, it is less than ideal. First of all, since the monochrome image and lightness channel are not identically the same, simply replacing them is insufficient. Secondly, the disparity between the two images still exists, interfering with accurate depth estimation. Additionally, since two cameras are used, it is worth further researching estimating depth in a stereo method. It will be future work to solve these problems.



\bibliographystyle{IEEEbib.bst}
\bibliography{root}

\end{document}